\def\year{2019}\relax
\documentclass[letterpaper]{article} %DO NOT CHANGE THIS
\usepackage{aaai19}  %Required
\usepackage{times}  %Required
\usepackage{helvet}  %Required
\usepackage{courier}  %Required
\usepackage{url}  %Required
\usepackage{graphicx}  %Required
\usepackage{amsmath} % Mingzhi added
\title{Investigating the Relationship between Multi-Party Linguistic Entrainment, Team Characteristics and the Perception of Team Social Outcomes}
\author{Mingzhi Yu,\textsuperscript{\rm 1}
Diane Litman,\textsuperscript{\rm 1}
Susannah Paletz\textsuperscript{\rm 2}\\
\textsuperscript{\rm 1}University of Pittsburgh,
\textsuperscript{\rm 2}University of Maryland\\
miy39@pitt.edu, dlitman@pitt.edu, paletz@umd.edu
}
\begin{document}
\maketitle
\begin{abstract}
Multi-party linguistic entrainment refers to the phenomenon that speakers tend to speak more similarly during conversation. We first developed new  measures of multi-party entrainment on features describing linguistic style, and then examined the relationship between entrainment and team characteristics in terms of  gender composition, team size, and diversity. Next, we predicted the perception of team social outcomes using multi-party linguistic entrainment and team characteristics with a hierarchical regression model. We found that teams with greater gender diversity had higher minimum convergence than teams with less gender diversity. Entrainment contributed significantly to predicting perceived team social outcomes both alone and controlling for team characteristics.
\end{abstract}

\section{Introduction}

Linguistic entrainment is the phenomenon that speakers tend to speak similarly in conversations \cite{brennan1996conceptual}. It has attracted attention for its potential to improve performance when being incorporated into spoken dialogue systems. One successful example is that \citeauthor{lopes2015rule} \shortcite{lopes2015rule} utilized entrainment to automatically choose better primes in the prompt of a dialogue system. In dyadic conversations, the existence of entrainment has been widely investigated for various linguistic features. \citeauthor{brennan1996conceptual} \shortcite{brennan1996conceptual} have found that speakers tend to refer to the same object using identical lexical terms in their conversation. \citeauthor{levitan2011measuring} \shortcite{levitan2011measuring} found evidence of prosodic entrainment in intensity, pitch and voice quality. \citeauthor{niederhoffer2002linguistic} \shortcite{niederhoffer2002linguistic} found that speaker pairs gradually matched their linguistic style in conversation. Besides linguistic features, researchers have also linked dyadic entrainment to other aspects of conversations, such as task success \cite{reitter2007predicting}, gender factors \cite{namy2002gender} and social behaviors \cite{levitan2012acoustic}. In recent years, researchers have started studying multi-party entrainment in conversations with more than two speakers. As with dyadic entrainment, the existence of multi-party entrainment has also been found for different linguistic features. For example, \citeauthor{rahimi2017entrainment} \shortcite{rahimi2017entrainment} found speakers in the same group entrained on high frequency words and topic words. \citeauthor{litman2016teams} \shortcite{litman2016teams} found  significant group-level differences in pitch, jitter and shimmer between first and second halves of conversation. Multi-party entrainment has also been associated with other aspects of conversations such as task performance and group cohesiveness. \citeauthor{friedberg2012lexical} \shortcite{friedberg2012lexical} found that higher scoring teams are more likely to entrain in the use of task-related terms. \citeauthor{gonzales2010language} \shortcite{gonzales2010language} suggested that group linguistic style matching is a significant indicator of team cohesiveness. Compared to studies demonstrating the existence of multi-party entrainment, however, fewer studies have investigated the links between multi-party entrainment and other aspects of conversations. We present two investigations of such relationships.

First, we investigate whether group or team characteristics relate to multi-party entrainment, since multiple individuals simultaneously engage in the same conversation. While dyad research has analyzed entrainment and  gender composition \cite{levitan2012acoustic,namy2002gender}, relationships between team characteristics and multi-party entrainment could be more complex, given the increasing number of person-to-person and person-to-team communications. We examine such relationships using cooperative game conversations, in which multiple speakers are brought together as a team. Three types of team characteristics are investigated:   gender composition as in prior work, as well as team size and diversity. 

Second, since relationships between multi-party entrainment and other conversational aspects are not well established, it is plausible that  correlations found in prior studies \cite{friedberg2012lexical,gonzales2010language} are confounded with team characteristics such as team size and gender composition. Team characteristics have already been found to have complex impacts on team processes \cite{o1989work,smith1994top,fisher2012facet}. For instance, team size is negatively correlated with team conflict \cite{amason1997effects}. We hypothesize that both entrainment and  team characteristics, specifically team size, gender composition and the diversity of a team, are associated with the perception of team social outcomes. We use hierarchical regression models to examine the unique contribution of multi-party entrainment in explaining perceived team social outcomes above and beyond  team characteristics.    

Finally, to support our studies, we have  developed an innovative representation of  multi-party entrainment by extending the measurement  from \citeauthor{litman2016teams} \shortcite{litman2016teams} and adapting it to study the feature of linguistic style from \citeauthor{pennebaker1999linguistic} \shortcite{pennebaker1999linguistic}. We used this measure to statistically examine  relationships between multi-party entrainment and team characteristics in a set of dialogues from  a freely available cooperative game corpus. We also demonstrated that  multi-party entrainment is  associated with team outcomes, even after controlling for team characteristics. 

\section{The Teams Corpus}\label{sec:data}
The freely available Teams Corpus \cite{litman2016teams} consists of 47 hours of audio and transcriptions from 62 teams (35 three-person, 27 four-person). The audio files are  manually segmented and transcribed at the level of inter-pausal units (IPUs), based on a pause length of 200 milliseconds. Each team consists of  American native speakers from 18 to 67 years old who played two rounds of the cooperative board game Forbidden Island.
213 individuals (79 males, 134 females)  were assigned to the teams and given one of  four game roles: Engineer, Messenger, Pilot, Explorer. 

The corpus also includes survey data. A pre-game survey collected personal information such as age, gender, and eight options for ethnicity. While each participant could choose multiple options, in this paper we categorize each speaker into nine exclusive categories: Caucasian (150), East Asian (12), South Asian (11), Pacific Islander (0), Black (15), Native American (0), Hispanic (3), Middle Eastern (2), and Multiple Ethnicity (20) for participants who chose more than one of the other categories. The gender data yields seven types of team gender composition: 0\% female (2), 25\% female (4), 33\% female (7), 50\% female (9), 66\% female (18), 75\% female (10),  100\% female (12). Participants also took post-game surveys to evaluate team processes. These surveys contained a series of self-report questions on team cohesion, satisfaction, and other team social outcome constructs.  

The studies presented in this paper are based only on  the data related to the first of the two games, as only these transcriptions  were available.
Before computing entrainment, we  further processed these transcripts by removing punctuation, converting all words to lower case, and removing a list of interjections, 
e.g., `hmm', that are not discussed in linguistic style \cite{pennebaker1999linguistic}. We then  concatenated all the  processed IPU transcriptions for each speaker.

\section{Features and Measures}
\subsection{Features of Speaker Linguistic Style}
Before computing entrainment, we first extracted linguistic style features for each speaker in each transcript using LIWC2007~\cite{pennebaker2007liwc2007}, a computational application for text analysis that includes a  dictionary mapping a list of words to  64 psychological and linguistic categories. We used this dictionary to label each word in each speaker's concatenated IPU transcripts with potentially multiple LIWC categories. The final number of occurrences of each category was then converted into a percent. 

In our study we only focused on a limited subset of LIWC categories, namely function words. The first reason is that function words reflect the speaker's psychological state and convey information about the interactive process \cite{chung2007psychological}. Function words represent a high-level linguistic difference in style. Second, in contrast to content words, function words do not rely on any specific task domain \cite{gonzales2010language} and have a very high frequency in daily speech \cite{rochon2000quantitative}. Using function words as features can alleviate feature sparsity. Since a considerable number of studies about linguistic style have used function words \cite{gonzales2010language,danescu2011mark,mukherjee2012analysis}, we directly adopted \citeauthor{gonzales2010language}'s \shortcite{gonzales2010language} selection of 9 LIWC categories as function words. 

\begin{figure}
\includegraphics[width=1\columnwidth]{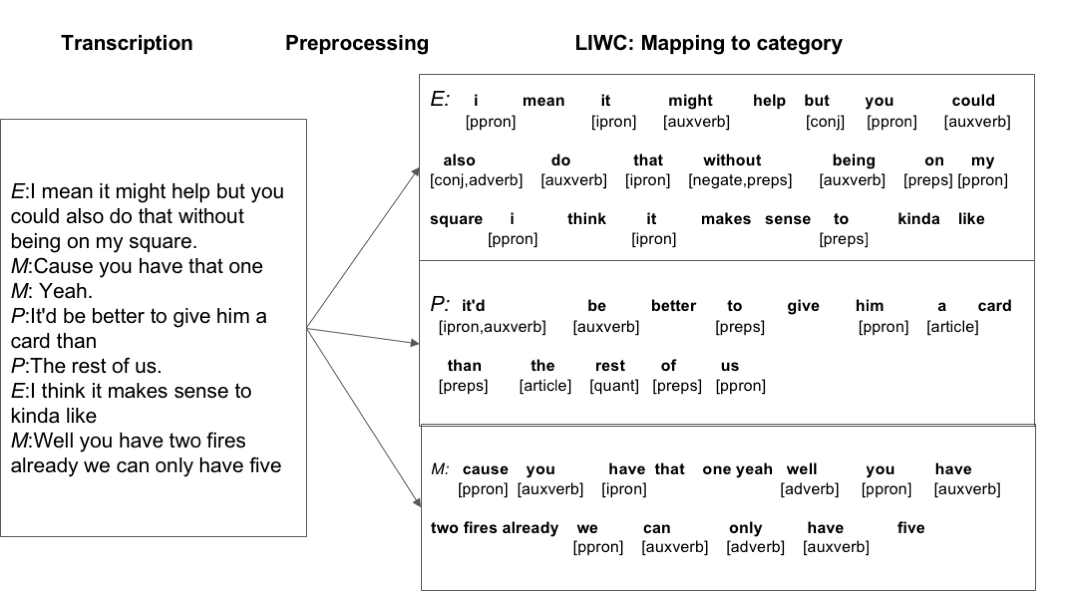}
\caption{Each tag corresponds to a LIWC function word category. negate: negation, conj: conjunctions, preps: prepositions, ppron: personal pronouns, ipron: impersonal pronous, article: article, adverb: adverbs, quant: quantifiers, auxverb: auxiliary verbs. }
\label{excerpt}
\end{figure}

Figure \ref{excerpt}  shows how we used LIWC to create function word features from a transcript excerpt.
After the transcript preprocessing and speaker IPU concatenation discussed above,  LIWC scored each speaker's input text and generated the category percentages for each of the 64 categories. For instance, the Engineer uttered 24 words in this excerpt but only one word belongs to the category negate. Thus, the category percentage for negate is 1/24 = 4.20\%. Since one word may belong to multiple categories, the sum of category percentages for the 64 categories may exceed 100.

\subsection{Measures of Team Linguistic Style Entrainment}
There are various methods to directly calculate multi-party entrainment using linguistic style. Some text-based studies have proposed probabilistic frameworks in linguistic style matching based on pairwise comparisons between  speakers \cite{mukherjee2012analysis,danescu2011mark}. However, compared to their data, our data has a lower density of reciprocated interactions per pair. The number of conversations between speakers is insufficient for constructing such probability models. Addressee identification to create appropriate pairs is also not straightforward. \citeauthor{gonzales2010language} \shortcite{gonzales2010language} developed a method to perform linguistic style matching based on multi-party speech, but they only focused on a global measure rather than on the degree of change. Recently, \citeauthor{litman2016teams} \shortcite{litman2016teams} proposed a method to compute multi-party entrainment on acoustic-prosodic features based on the same Teams Corpus as used here. Their method highlighted  feature change  over time, which is more relevant to linguistic style entrainment.
 
For each feature, they calculated the  difference between a pair of speakers as the absolute difference of feature values, and the team difference as the average  difference over all pairs. In our study, linguistic style is a single feature with multiple categories, so we converted their calculation of pair differences by summing up all the category differences. Moreover, we weighted category differences by the frequency of categories. More specifically,  $TDiff_{unw}$ (unweighted team difference) converts the team difference of \citeauthor{litman2016teams} \shortcite{litman2016teams} to deal with multiple feature categories. $TDiff_{w}$ (weighted team difference) extends $TDiff_{unw}$  by weighting the category differences similarly to \citeauthor{gonzales2010language} \shortcite{gonzales2010language}. We calculated both $TDiff_{unw}$ and $TDiff_{w}$  for each pair of speakers and then averaged over all pairs. The formulas are shown in Equations \ref{TDiff unw}, \ref{TDiff w}, and \ref{KDiff}, where $F, K$, and $|team\ size|$ respectively refer to the function word category set, an arbitrary function word category, and the team size.  $KDiff_{ij}$ refers to the weighted category difference of category K between speakers i and j.

\begin{equation}
\label{TDiff unw}
TDiff_{unw} = \frac{\sum_{\forall i \neq j \in team}(\sum_{K \in F}(|K_i-K_j|)}{|team\ size| * (|team\ size| - 1)}
\end{equation}

\begin{equation}
\label{TDiff w}
TDiff_{w} = \frac{\sum_{\forall i \neq j \in team}(\sum_{K \in F}(|KDiff_{ij}|)}{|team\ size| * (|team\ size| - 1)}
\end{equation}

\begin{equation}
\label{KDiff}
KDiff_{ij} = \frac{|K_i-K_j|}{K_i+K_j}, KDiff_{ij} = 0  \text{ if }K_{i}, K_{j} = 0
\end{equation}

 \citeauthor{litman2016teams} \shortcite{litman2016teams} then define convergence, a type of entrainment measuring increase in feature similarity, by comparing the $TDiff$ of two non-overlapping temporal intervals of a game as in Equation \ref{convergence}. $C_{ij}$ and $TDiff$ refer to the team's convergence and the weighted (or unweighted) team differences, respectively. Assuming the game is divided into n disjoint temporal intervals, i and j  refer to two predetermined temporal intervals in chronological order. 
\begin{equation}
\label{convergence}
 C_{ij} = TDiff_i-TDiff_j, i < j \in n
\end{equation}
However, this definition leaves two unanswered questions. First, the measure of convergence allows negative values that represent divergence, which is the tendency that team members speak differently. Second, it requires the researcher to hand pick temporal intervals that are not guaranteed to result in an optimal measurement of entrainment. Hence, we derived four new variables of convergence (see Equations \ref{Max and Min} and \ref{absMax and absMin}): Max and Min calculating the maximum and minimum positive $C_{ij}$, and absMax and absMin calculating the absolute maximum and minimum $|C_{ij}|$. 

\begin{equation}
\label{Max and Min}
Max \text{ or } Min= Max\{C_{ij} > 0 \} \text{ or } Min\{C_{ij} > 0 \}
\end{equation}

\begin{equation}
\label{absMax and absMin}
absMax \text{ or } absMin = Max \{|C_{ij}|\} \text{ or } Min\{|C_{ij}|\}
\end{equation}
Rather than two fixed intervals, we iterated over all two arbitrary temporal intervals in chronological order and conducted the comparison.
Consequently, the Max and Min only measure maximum and minimum convergence so that they directly reflect the decrement of $TDiff$ between two intervals. The absMax and absMin measure the maximum and minimum magnitude of the change of $TDiff$ in the entire conversation. Unlike the Min and Max, the absMax and absMin are determined by the values of convergence or divergence. We added the absMax and absMin beyond Min and Max so that they reflect the overall fluctuation ranges of $TDiff$, which might also be an important aspect of entrainment. Therefore in total, we defined eight measures of team entrainment: {\bf unweighted} and {\bf weighted Max, Min, absMin}, and {\bf absMax convergence}.  

The parameter n in Equation~\ref{convergence} determines the length of temporal intervals being compared. Many studies defined n as two so that the conversation is evenly divided into two halves \cite{levitan2011measuring,rahimi2017entrainment}. Since \citeauthor{litman2016teams} \shortcite{litman2016teams}  previously found that in the Teams corpus the highest acoustic-prosodic convergence occurred within the first and last three minutes, we used this finding to define our n. We evenly divided each game, which was limited to 30 minutes, into ten intervals, so each interval is less than three minutes. Since our focus is on measure development in this paper, 
methods for optimally tuning  this temporal parameter are left for future work.

\begin{figure}[!ht]
    \centering
    \includegraphics[width=1\columnwidth]{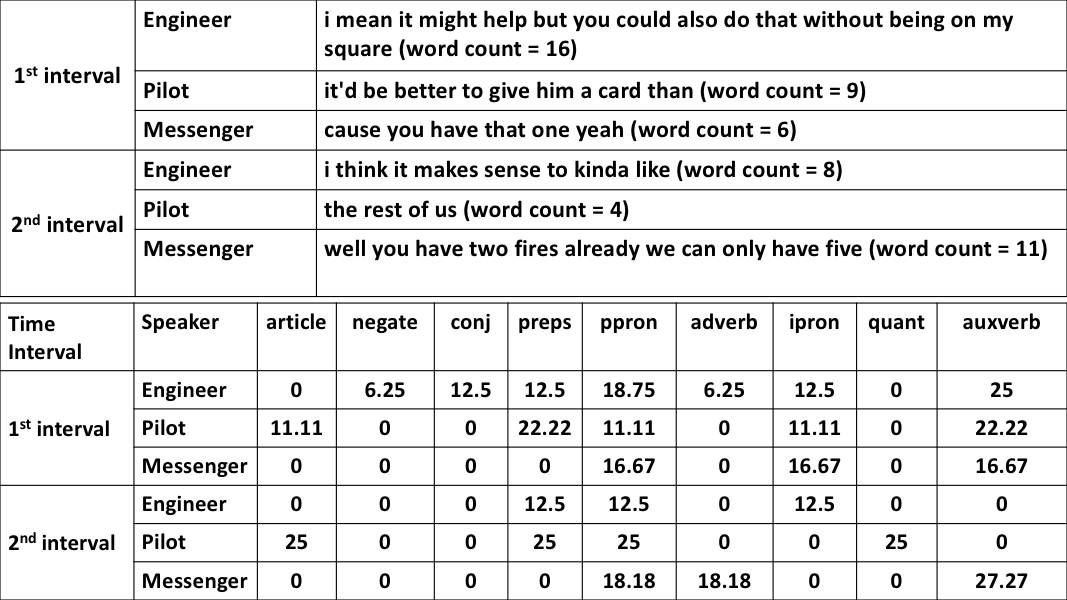}
    \caption{ Top: The input text of each speaker per interval. Bottom: The corresponding LIWC category percentage.}
    \label{fig:excerpt example}
\end{figure}

We will use Figure \ref{excerpt}'s excerpt to illustrate our calculations. Assuming n is set to two, we first divide the excerpt  into two time intervals. Assuming 
that the temporal midpoint of the excerpt occurs after the fourth IPU, the first interval includes the first through fourth IPUs. The second interval includes the fifth through seventh IPUs. For each speaker, all IPUs in each interval are concatenated and input to LIWC. The interval division and LIWC category percentage output are shown in Figure \ref{fig:excerpt example}. Based on Equation \ref{TDiff unw}, the unweighted pair difference between the Engineer and Pilot in the first interval is calculated as the sum of the absolute differences of all categories, which is equivalent to $ |0-11.11| + |6.25-0| + |12.5-0| + |12.5 - 22.22| + |18.75-11.11| + |6.25-0| + |12.5-11.1| + | 0 - 0| + |25-22.22|  = 57.64 $. Similarly, the pair differences between the other two pairs (Engineer and Messenger, Pilot and Messenger) are 52.08 and 50. The unweighted team difference is the average of these pair differences, which is 53.24.
The weighted team difference is calculated using Equation~\ref{TDiff w}, with the pair difference now being normalized by the frequency of each category. For instance, the absolute difference between Engineer and Pilot of negate is $|6.25 - 0| = 6.25$. This number is less than the absolute difference of $|18.75-11.11|=7.64$ for the category ppron.  However, the occurrence of negate is less common than ppron in the speech of Engineer and Pilot. The weighted difference of negate is $|6.25 - 0|/(6.25 + 0) = 1$, which is now greater than the weighted difference of ppron which is  $|18.75-11.11|/(18.75+11.11) = 0.26$. 
 
\subsection{Measures of Team Characteristics}
This paper focuses on the following team characteristics: {\bf team size}, 
{\bf gender diversity} (Blau's index and female percentage), {\bf ethnic diversity}, and {\bf age diversity}. Note that the female percentage measures the numerical female dominance in a team, while gender diversity indicates the variability of gender composition. Diversity of age, which has continuous values, is measured by the population standard deviation. Diversity of ethnicity and gender with categorical values is measured by Blau's index of heterogeneity \cite{blau1977inequality} as in Equation \ref{blau}, where $P_k$ is the proportion of a specific category k. 

\begin{equation}
Ethnic/Gender\ Diversity_{\ Blau's} = 1 - \sum{P_k}^2
\label{blau}
\end{equation}

\subsection{Measures of Perceived Team Social Outcomes}
We assessed the perception of team social outcomes using the existing self-reported post-game survey responses. The survey contains scales related to team processes and team conflict. Team processes consist of the perceptions of team cohesion, general team satisfaction, potency/efficacy, and perceptions of shared cognition \cite{wendt2009leadership,wageman2005team,guzzo1993potency,gevers2006meeting}. These four measures were strongly correlated with each other. Thus, we aggregated them into a single scale by averaging their z-scored scale composites, Cronbach's $\alpha$ = 0.78. Team conflict consists of task, process and relationship conflict. These three types of conflict reflect the topic of the conflict, be it about the task at hand, work processes, or interpersonal values and personal relationships. Process conflict is consistently negatively related to performance, but task conflict is positively related to performance under some conditions \cite{de2012paradox}. Therefore, we kept these three types of conflict as individual variables. Overall, we thus have four measures of perceived team social outcomes: {\bf team processes, task conflict, process conflict} and {\bf relationship conflict}.

\section{Results}

\subsection{Relating Team Characteristics and  Entrainment}
We first tested the relationship between linguistic style entrainment and team characteristics with continuous values (gender, ethnic and age diversity) using Spearman rho correlations. There was a significant positive correlation between unweighted convergence Min and gender diversity, $(r(62)=.22, p<.05)$. This correlation indicated that teams with greater gender diversity had higher minimum convergence than teams with less gender diversity. 

We then performed one-way ANOVA tests between linguistic style entrainment and the categorical team characteristics, i.e.,  percentage of females and  team size. The unweighted absMax was found to significantly vary with female percentage for the 7 conditions (see corpus section), F(6,55) = 2.79, p = .019. Tukey HSD post hoc tests indicated that the 25\% condition (N = 4, M = 40.15, SD = 13.263) was significantly different with the 50\% condition (N = 9, M = 19.56, SD = 9.435), 66\% condition (N = 18, M = 19.39, SD = 9.407) and 75\% condition (N = 10, M = 18.92, SD = 8.117). The mean of the 25\% condition was larger than all other three conditions. This finding suggests that the maximum magnitude of the change of unweighted team differences in the 4-person team with one female was greater than other mixed-gender teams with more than one female.

\subsection{Predicting Perceived Team Social Outcomes}
We predicted four measures of perceived team social outcomes: team processes (MIN = -2.57, MAX = 1.51, M = 0.00, SD = 0.80); task conflict (MIN = 1.00, MAX = 3.33, M = 1.75, SD = 0.46); process conflict (MIN = 1.00, MAX = 3.00, M = 1.58, SD = 0.41) and relationship conflict (MIN = 1.00, MAX = 1.75, M = 1.15, SD = 0.20). A hierarchical linear regression (HLR) model allows us to consider the impact of team characteristics on the perception of team social outcomes, and then examine the significance of multi-party entrainment as a predictor beyond or controlling for team characteristics. Two models, Model 1 (M1) and Model 2 (M2), were constructed. Team size and team diversity (gender, ethnic and age) were entered simultaneously as independent team characteristic variables (IVs). Multi-party entrainment was entered into M2 as an IV beyond the team characteristics. Only variables of multi-party entrainment that significantly contributed to the model were selected in M2. The dependent variable (DV) of each HLR was the variable describing the perceived team social outcomes.

\begin{table}[t]
\centering
\resizebox{.95\columnwidth}{!}{
\begin{tabular}{lllll}
\hline
\textbf{DV}      &       & \textbf{IV}         & \textbf{Model 1 ($\beta$)} & \textbf{Model 2 ($\beta$)} \\
\hline
        &       & Age        & 0.04              & -0.03             \\
        &       & Ethnic     & 0.19              & 0.19              \\
        &       & Gender     & -0.04             & -0.11             \\
Task    &       & \%Female   & -0.21             & -0.30             \\
        &       & Team size  & 0.24              & 0.25*             \\
        &       & unw absMax &                   & -0.31*            \\
        \cline{3-5}
        & $R^2$ &            & 0.13              & 0.22              \\
        & F     &            & 1.60              & 2.54*             \\
\hline
        &       & Age        & 0.06              & 0.04              \\
        &       & Ethnic     & 0.25              & 0.29*             \\
        &       & Gender     & -0.08             & -0.13             \\
Process &       & \%Female   & -0.08             & -0.14             \\
        &       & Team size  & 0.34**            & 0.31*             \\
        &       & w absMax   &                   & -0.36**           \\
        \cline{3-5}
        & $R^2$ &            & 0.16              & 0.28              \\
        & F     &            & 2.10              & 3.57**            \\
\hline
\end{tabular}}
\caption{Predicting  team outcomes using hierarchical regression. Age: $Age Diversity_{SD}$, Ethnic: $Ethnicity Diversity_{Blau's}$, Gender: $Gender Diversity_{Blau's}$, \% Female: Percentage of female, w absMax: weighted convergence absMax, unw absMax:  unweighted convergence absMax. $\beta$: standardized Beta. * if p $<$0.05, ** if p $<$0.01.}
\label{Regression}
\end{table}
 
Significant HLR models are shown in Table \ref{Regression}. The M2 predicting the task and process conflict were both significant, but no M1 was significant. In the HLR predicting task conflict, no team characteristics contributed significantly to M1. Introducing variables of multi-party entrainment to M2 explained an additional 9.2\% variation in task conflict and the $\Delta R^2$ was significant, $\Delta F (1,55) = 6.46, p < 0.05$. Unweighted absMax and team size were both significant contributors to task conflict in M2. The negative association between unweighted absMax and task conflict suggested that the higher maximum magnitude of the change of team difference signaled less team conflict. Meanwhile, the positive association between team size and task conflict in M2 added evidence to previous findings that team size is positively associated with team conflict \cite{amason1997effects,smith1994top}.

In the HLR predicting process conflict, team size contributed significantly to M1. Adding multi-party entrainment (the weighted absMax) to M2 explained an additional 12.2\% of the variability in process conflict, and the $\Delta R^2$ was significant, $\Delta F (1,55) = 9.36, p < 0.01$. Multi-party entrainment along with team size and ethnic diversity were important predictors to M2. Team size and ethnic diversity were both positively associated with process conflict. We observed a negative association between the weighted absMax and process conflict. This finding implied that the higher maximum magnitude of the change of team difference signaled less process conflict. 

Overall, we found a negative association between maximum magnitude of the change of team difference and team conflict, specifically process and task conflict. Team size and ethnic diversity both had effects on team conflict. Maximum magnitude of the change of team difference was a significant predictor in team conflict. 

To determine whether the team characteristics had a significant impact on the conflict variables above and beyond the effect for entrainment, we switched the IVs in M1 and M2. Variables of entrainment were entered into M1 stepwise and then the team characteristics that had shown significance in the previous HLR were entered into M2 (see Table \ref{HLR: reverse}). We observed similar findings in that both M1 and M2 significantly predicted task and process conflict. The maximum magnitude of the change of the team difference was significantly negatively associated with task and process conflict. Team size and, for process conflict, ethnic diversity were significantly related to conflict above and beyond entrainment.

\begin{table}[t]
\centering
\resizebox{.95\columnwidth}{!}{
\begin{tabular}{lllll}
\hline
\textbf{DV}      &       & \textbf{IV}         & \textbf{Model 1 ($\beta$)} & \textbf{Model 2 ($\beta$)} \\
\hline
        &       & unw absMax & -0.27*            & -0.27*            \\
Task    &       & Team size  &                   & 0.25*             \\
        \cline{3-5}
        & $R^2$ &            & 0.07              & 0.13              \\
        & F     &            & 4.77*             & 4.55*             \\
\hline
        &       & w absMax   & -0.34**           & -0.34**           \\
Process &       & Team size  &                   & 0.32**            \\
        &       & Ethnic     &                   & 0.26*             \\
        \cline{3-5}
        & $R^2$ &            & 0.12              & 0.26              \\
        & F     &            & 7.87**            & 6.69**            \\
\hline
\end{tabular}}
\caption{Flipped HLR : Entrainment was stepwise entered in M1. Team characteristics showing significance in prior HLRs were entered in M2.}
\label{HLR: reverse}
\end{table}

\section{Conclusions and Future Work}
We first proposed a new method for measuring  multi-party linguistic style entrainment by converting and extending methods developed in prior studies of both linguistic style matching and team acoustic-prosodic entrainment. We then examined the relationship between multi-party entrainment and team characteristics. Our analysis implies that teams with greater gender diversity had greater minimum convergence than teams with less gender diversity, similarly to the findings of \citeauthor{levitan2012acoustic} \shortcite{levitan2012acoustic} and \citeauthor{namy2002gender} \shortcite{namy2002gender} that mixed-gender pairs generally entrain more in dyadic conversations. Moreover, the 4-person teams with more than one female had a higher maximum magnitude of change in team difference. Perhaps the existence of a female subgroup reconciled the team difference in these teams. In conclusion, different gender compositions affect the entraining behaviors of the overall team. These findings show that gender plays an important role for linguistic entrainment in human interactions. They also reveal a need to study the underlying process of multi-party entrainment with different granularity levels. 
Next, we predicted the perception of team social outcomes by team characteristics and variables of entrainment with hierarchical regression models. The experimental results indicated that the maximum magnitude of the change of the team difference was negatively associated with team conflict. Adding this variable of entrainment beyond team characteristics resulted in  statistically significant improvements in model prediction. Finally, by entering entrainment variables in the first rather than second model, we showed that entrainment was significantly negatively associated with task and process conflict, both when controlling for team characteristics and when not. Although the overall models did not account for a large amount of variance, the base model of only team characteristics was improved significantly by adding entrainment.  In sum, we found that entrainment is a promising feature to predict team social outcomes. In terms of broader impact, we can now possibly evaluate the success of team conversations using linguistic style entrainment. Additional interdisciplinary research building on our findings could test whether entrainment mediates the effects of team characteristics on social and task outcomes in different settings.

In future work, we will investigate different feature combinations and prediction models to improve the performance of our statistical models. To further improve the calculation of multi-party entrainment, we intend to search for an optimal temporal window. Additionally, we plan to review the validity and accuracy of the team social outcomes, which were measured with self-reported surveys. We also plan to investigate the relationship between multi-party entrainment and individual differences, such as personality and education background.

\section{Acknowledgements}
This work is supported by the National Science Foundation under Grant Nos. 1420784 and 1420377. We thank all the transcribers, the Pitt PETAL group and the anonymous reviewers for advice in improving this paper.  

\bibliographystyle{aaai}
\bibliography{flair2018}

\end{document}